\documentclass[conference]{IEEEtran}
\IEEEoverridecommandlockouts

\usepackage{cite}
\usepackage{amsmath,amssymb,amsfonts}
\usepackage{fancyhdr}
\usepackage{multirow}
\usepackage{array}
\usepackage{setspace}
\usepackage[ruled,linesnumbered]{algorithm2e}
\usepackage{soul}
\usepackage[dvipsnames]{xcolor}
\usepackage{bm}
\usepackage{hyperref}
\usepackage{cleveref}
\usepackage{makecell}
\usepackage{graphicx}
\usepackage{mathtools}
\usepackage{comment}
\usepackage{authblk}

\makeatletter
\def\blfootnote{\xdef\@thefnmark{}\@footnotetext}
\makeatother

\def\BibTeX{{\rm B\kern-.05em{\sc i\kern-.025em b}\kern-.08em
    T\kern-.1667em\lower.7ex\hbox{E}\kern-.125emX}}
\newif\ifmaximal
\maximaltrue 

\begin{document}
\ifmaximal
\title{Computation on Sparse Neural Networks: an Inspiration for Future Hardware}
\fi
\author{Fei Sun\textsuperscript{1}\thanks{\textsuperscript{1} Email: \href{f.sun@alibaba-inc.com}{f.sun@alibaba-inc.com}.}}
\author{Minghai Qin}
\author{Tianyun Zhang}
\author{Liu Liu}
\author{Yen-Kuang Chen}
\author{Yuan Xie}
\affil{Alibaba DAMO Academy}

\maketitle

\begin{abstract}
Neural network models are widely used in solving many challenging problems, such as computer vision, personalized recommendation, and natural language processing. Those models are very computationally intensive and reach the hardware limit of the existing server and IoT devices. Thus, finding better model architectures with much less amount of computation while maximally preserving the accuracy is a popular research topic. Among various mechanisms that aim to reduce the computation complexity, identifying the zero values in the model weights and in the activations to avoid computing them is a promising direction. 

In this paper, we summarize the current status of the research on the computation of sparse neural networks, from the perspective of the sparse algorithms, the software frameworks, and the hardware accelerations. 
We observe that the search for the sparse structure can be a general methodology for high-quality model explorations, in addition to a strategy for high-efficiency model execution.
We discuss the model accuracy influenced by the number of weight parameters and the structure of the model. The corresponding models are called to be located in the {\em weight dominated} and {\em structure dominated} regions, respectively. We show that for practically complicated problems, it is more beneficial to search large and sparse models in the weight dominated region. 
In order to achieve the goal, new approaches are required to search for proper sparse structures, and new sparse training hardware needs to be developed to facilitate fast iterations of sparse models. 

\end{abstract}

\section{introduction}

In the past few years, artificial intelligence (AI) have significantly changed our lives. New technologies such as autonomous driving, personalized recommendation, and language translation are all backed by complicated neural network models. The advancement of AI is mainly driven by three factors: the algorithm innovation, the amount of data, and the computing power~\cite{aiandcompute}. 
\ifmaximal
Among the three, the computing power supports both searching for larger and better models, and processing enormous amount of data.
\fi

\ifmaximal
With the advancement of the algorithms and the amount of computation, problems previously considered challenging now perceived as trivial. And newer problems requiring much higher computing power seem to be within the reach of hands. 
\fi

However, the amount of computing resource is limited. Several methods have been proposed to reduce the computation of the AI tasks, in particular, the deep learning approach. One of the most important concepts in deep learning based AI is a {\em tensor}, which can be seen as a multi-dimensional matrix. Since the computation of  tensors grows dramatically with the tensor size, there is a variety of methods to decompose a tensor into several low-rank tensors. Polyadic tensor decomposition~\ifmaximal\cite{Polyadic1927JMP,Pwaytensor1928JMP}\fi~is proposed to approximate a $d$-way tensor into $d$ rank-one tensors. 
\ifmaximal
Tucker decomposition~\cite{Tuck1963a} is proposed as a higher-order principal component analysis (PCA) to decompose a tensor into a core tensor multiplied (or transformed) by a matrix along each dimension. Other variants~\cite{Carroll1970AnalysisOI,parafac1972ucla,candelinc1980,Harshman96uniquenessproof} impose constraints such as linearity or symmetry onto Polyadic or Tucker decomposition.
\fi
Some modern neural network architectures, such as MobileNetV2~\cite{mobileNetv22018cvpr}, take advantages of the decomposition of 4-D convolutional kernels. More recently, Winograd convolution is proposed in~\ifmaximal~\cite{winograd1980IAM,winofft2016cvpr}\fi~to reduce the arithmetic complexity.
\ifmaximal
~\cite{FCNN2017ecml, AcceleratingCN2018VLSI} use fast Fourier transform to accelerate convolution neural networks. 
\fi
Another approach to reduce the computation of deep neural networks is to quantize the model, {\it i.e.}
use less number of bits to represent weights and activation~\cite{quantizeDNN2017MLR}. Similarly, after a pioneering study on the neural network compression~\cite{Han2016a}, it has become a hot topic to improve the performance of neural networks when a majority of the parameters are set to zero to form a {\em sparse} model such that most of the computation can be avoided through carefully designed software and hardware implementations. 

\ifmaximal
In this paper, we survey the recent research progress on sparsity. From the algorithm side, we survey the static sparsity, the dynamic sparsity, and the sparsely activated models. On the software side, we describe different sparse representations, and compare some sparse libraries. On the hardware side, we examine the existing hardware accelerators for training and inference. 

We also describe several complicated problems that push the current computation to its limit, and explain that sparse computation may be a viable solution.
\fi
In order to achieve that, we introduce the concept of {\it weight dominated} and {\it structure dominated} regions as the model size increases. We illustrate that in the weight dominated region, sparse models are more accurate than the corresponding dense models with the same number of floating point operations (FLOPs). However, it is not possible to apply the conventional pruning algorithms to large and dense models to create the complicated sparse models close to the computing limit, and therefore the sparse models need to be derived from the smaller dense models. Thus, the approach to find a reasonable sparse structure is also a model exploration problem. 

In addition to the necessary algorithm innovations to create large and sparse models close to the computing limit, we also need to invest in the  sparse training platforms and the corresponding training software frameworks, so that the entire training procedure can be done efficiently within the computing limit. 

\ifmaximal
We organize this paper as follows. In Section~\ref{sec:complicated_problems}, we describe a few complicated problems that may benefit from the computation on sparse models, followed by a survey of the existing pruning algorithms in Section~\ref{sec:sparse}. In Section~\ref{sec:solutions_to_complicated_problems}, we outline our reasons that sparsity may be a viable solution to those complicated problems. We survey the sparse software and hardware solutions in Sections~\ref{sec:sparse_software} and~\ref{sec:sparse_hardware}, respectively. Then we identify some future research directions on the sparse algorithms and those enabled by the sparse training hardware in Section~\ref{sec:research_directions}. We conclude in Section~\ref{sec:conclusion}.
\fi

\ifmaximal
\section{The Challenges on Solving Real-life Complicated Problems} \label{sec:complicated_problems}

The introduction of the neural networks fundamentally changes researchers' approach to solve complicated problems. With sufficient data, once a \ifmaximal promising \fi neural network model architecture is identified, the solution is to find the most complicated model that is economically viable and practically usable.

\ifmaximal
In 2012, the image classification was considered a complicated problem, as the model quality at the time was far from being acceptable. Training an AlexNet on ImageNet dataset took five to six days on a two-GPU machine~\cite{Krizhevsky2012}. With the recent improvements in computing power, training a ResNet-50 on ImageNet dataset can be shorten to two minutes~\cite{mikami2018sgd}. Supported by the computing power, many neural network models have surpassed human-level accuracy~\cite{he2015rectifiers} and thus the single image classification problem is no longer considered a complicated problem. 

However, this is just the beginning of artificial intelligence.
Many new problems limited by the computing resources are still considered complicated. Below are a few examples.
\else
Below are a few problems that are limited by the computing resources and are considered complicated problems.
\fi

Tasks such as video classification or action recognition are widely used in autonomous driving \cite{makantasis2015deep} and game playing \cite{mnih2013playing}. With an additional time dimension, the amount of data to process is increased significantly. Limited by the computing power, the requirement of real-time latency also hinders the feasibility of deploying large models for video analysis.

Similarly, the 3D tasks based on point cloud data, such as semantic segmentation and classification, raise challenges in both computation and model architecture exploration. Due to the discrete and sparse nature of the inputs, a recent model named PointNet++~\cite{qi2017pointnet++} feeds the embedded inputs to a hierarchical multi-layer perceptron (MLP); and MinkowskiEngine~\cite{choy20194d} proposes a generalized sparse convolution. With  billions of points in the dataset, the speed of processing them is usually the bottleneck. 

Graph neural network (GNN) models~\cite{scarselli2008graph}  are emerging rapidly to be used in social networks~\cite{degenne1999introducing} and risk control~\cite{summala1988risk}. The sparse nature of the inputs with billions of nodes and edges puts the existing training and inference systems under enormous stress.
Being both computation and bandwidth bounded depending on the phase, the model training suffers from sub-linear scaling on massive parallel machines. 

Self-attention based transformer model architecture is widely used in natural language processing (NLP) tasks. It is known that the prediction accuracy increases on larger models, and the number of weights of the largest model, GPT-2 8B~\cite{radford2019language}, exceeds 8 billion, which requires 512 GPUs to train. 

Since 2012, the amount of computation used to train the largest models doubles every 3.4 month~\cite{aiandcompute}. The trend of computation scaling is likely to continue technologically, but at a much higher price tag. For example, a single training run of XLNet model on Google cloud TPU v3 costs as much as \$60K~\cite{trainingcost}. Thus, the searching for better models is only affordable by affluent institutions on high rewarding models. 

The trend to endlessly continue scaling the computation is less likely to continue economically. Thus we believe that searching for highly effective and efficient models using alternative methods is prominent. Among various approaches, intentionally inducing zeros to the models and avoiding computing them is a promising direction. 
\fi

\section{Sparse Models and Sparse Computation} \label{sec:sparse}
In this section, we survey the status of research on sparse neural network model computation.
The sparsity of the neural network execution can be generally categorized into two types, {\em static} sparsity and {\em dynamic} sparsity. Static sparsity refers to the reduction of non-zero weights in neural network models. Once the positions and values of the non-zero weights are determined, they are fixed in inference. Therefore, the amount of computation is constant for different inputs.  On the other hand, dynamic sparsity reduces the computation within a neural network layer {\em dynamically} based on the computational characteristics during inference or training. 
The recent development of large deep learning models enables them to be {\em sparsely activated} for each input examples and for each learning task.

\subsection{Static Sparsity} \label{sec:static_sparsity}

\ifmaximal
One of the approaches to increase the representation capability of a neural network is to increase the model size, usually evaluated by the number of weights (connections between neurons). 
Therefore, modern neural network designs have gone through a period of time when the number of weights is increased drastically. On the other hand, the intrinsic redundancy within large models implies that the non-critical weights can be identified and pruned with minimum accuracy loss compared to the dense models. 
\fi
The training process to obtain a pruned neural network in the existing literature usually follows three steps. First, a large and dense neural network is pre-trained. Second, in the pruning process, some non-critical weights are identified and set to zero permanently. Third, the relatively sparse neural network is re-trained and expected to obtain similar accuracy to the dense one. 
The three steps can be applied repeatedly to gradually increase the sparsity and maintain an acceptable accuracy.
This three-step pruning process requires to pre-train a large and dense model in the first step as a superset of the sparse model. In Section~\ref{sec:solutions_to_complicated_problems}, we will point out that this approach is infeasible to obtain large and sparse models for complicated problems. 

Based on the location characteristics of the remaining non-zero weights, static sparsity can be achieved via {\em irregular} pruning or {\em structured} pruning.
\ifmaximal
\fi

Irregular pruning aims to remove non-critical weights without any constraints on their locations, the key of which is to define the importance of each weight.~\cite{han2015learning} proposes a magnitude-based importance metric and iteratively prune weights with small magnitude. After that, the pruning ratio is largely improved by different approximation and optimization methods.~\cite{LearningL02017iclr} uses $\ell_0$ regularization and  ~\cite{wen2016learning} manages to solve the non-convexity problem of  $\ell_0$ by $\ell_1$ approximation.~\cite{zhang2018systematic,ren2019admm} utilize an effective technique in optimization theory to improve the sparsity based on $\ell_0$ regularization. 
\ifmaximal
Other interesting works
\cite{lotteryTicket2018iclr,Rethinking2019ICLR} 
explore the feasibility of re-training the sparse neural networks without a pre-trained model. 
\fi

\ifmaximal
While irregular pruning provides much flexibility of zeroing the weights and thus maximally preserves the accuracy, the overhead of representing the locations of non-zero weights and interpreting them during inference cannot be overlooked. For example, compressed sparse row (CSR) and compressed sparse column (CSC)~\cite{CSR1967IEEE} are two popular sparse matrix formats but they all require the non-zero elements to be extracted sequentially. This limits the throughput of non-zero weights to be read out and calculated.
\fi

On the other hand, structured pruning imposes constraints on the locations of non-zero weights and thus reduces the irregularity. 
It improves the execution efficiency of the sparse models on existing devices, but often reduces the sparsity level in order to maintain the same accuracy.
Different structures are proposed for different computing devices. For example, filter-wise pruning and shape-wise pruning~\cite{wen2016learning} can remove rows and columns in matrix-matrix multiplications, which translates to computation reduction in GPU platforms. Recent works explore some special dimensions and propose pattern pruning and kernel-wise pruning~\cite{ma2019pconv}. These structured prunings have finer scales than filter- and shape-wise pruning but find themselves useful in edge computing devices such as mobile platform.

Depending on the structure of the pre-trained dense model, the sparsity level of the pruned model can reach 90-97\% on some heavily over-parameterized models ({\it e.g.} VGG, AlexNet, RNN models), and 50-70\% on some compact models ({\it e.g.} MobileNetV1/V2/V3).

Both irregular and structured pruning are performed during the training phase, which is before the deployment of the neural networks. Therefore, the amount of computation for the sparse neural network models can be accurately predicted with little variations. 
\ifmaximal
However, this benefit comes with a drawback, that is, the neural network models do not differentiate inputs that are difficult to recognize from the easy ones that  intuitively can be correctly recognized with less computation.
\fi

\subsection{Dynamic sparsity}
Dynamic sparsity refers to neural network models and computation mechanisms that adjust the amount of computation  according to different input data and internal signals inside the neural networks during inference. 
\ifmaximal
So far, its sole purpose is to improve the computation efficiency without affecting the model performance. 
\fi

During dynamic inference, computation is reduced by pruning the activation in the neural networks~\cite{ActvationPrune2017GlobalSIP}, by skipping the calculation of zeros after the $ReLU$ function~\ifmaximal\cite{ZeNA2017DandT,parashar2017scnn,FPGAzeroSkipping2018ISRSP}\fi~
or by dynamically change the number of bits in quantization~\cite{EEquantize2018isca}. Besides, $ReLU$-induced sparsity prediction methods in convolutional neural networks (CNNs) have been proposed to skip computations dynamically~\ifmaximal\cite{dong2017complicated,gao2018dynamic,cao2019seernet,hua2019channel}\fi. 

Many mechanisms to explore irregular sparsity in dynamic inference rely on the hardware implementation. Other than $ReLU$-based dynamic computation skipping, the special cell structure and the temporal input similarity have enabled computation and update skipping in recurrent neural networks (RNNs)~\ifmaximal\cite{neil2017delta,zhang2018towards,campos2018skip}\fi.

Note that the difficulty of a learning task can depend on the input examples. 
The computation of neural networks can be reduced if an easy input can be identified and use a much simpler data and network flow. \ifmaximal\cite{Teer16icpr,MultiScaleDN2017iclr,SacrificingAF2018icann,SpatiallyAC2017cvpr,shallowdeepnetworks2019icml}\fi~
propose to output the final learning decision by early exiting if the confidence score is above a threshold inside the neural networks.~\cite{AdaptiveNN2017icml} creates a directed acyclic graph where each node is a pre-trained deep neural network (DNN) with simpler DNNs at the source and complex DNNs at the sink. Then an exit policy is trained to determine which DNN to go through.\ifmaximal\cite{CNNadaptiveGraph2018eccv,SkipNet2018ECCV,blockdrop2018cvpr}\fi~
propose to skip part of the sequential layers based on the dynamics of some gates in the networks, the decisions of which are trained by the Gumbel sampling~\cite{Gambel2017iclr} or the reinforcement learning. Another mechanism to use a gate to guide the neural network to execute one of several parallel branches is proposed in~\cite{Mullapudi2018CVPR}. Besides dynamic inference, dynamic sparse graph is proposed to reduce computational and representational costs at DNN training~\cite{liu2018dynamic}. Recently,~\cite{onceForAll2020iclr} proposes an efficient progressive shrinking method to train a super-network based on some architecture and deploy only a portion of it, which outperform  the base architecture (and its NAS extensions) such as EfficientNet or MobileNetV3.

The sparsity of the activation can reach 50-70\% post $ReLU$. Much higher sparsity can be reached if larger structures or even layers are skipped at the cost of some accuracy loss. 

In some most complicated problems, such as graphs and point clouds, the inputs are inherently sparse. 
This sparsity is currently not sufficiently taken advantages of to explore the model structure, and we believe this may be an interesting area of research.


\subsection{Sparsely activated models}
Single-task single-model deep learning suffers from the small model capacity to handle complicated problems.~\cite{sparseMoE2017iclr} proposes a sparsely-gated mixture-of-experts (MoE) layer to enable only two or three sub-models (out of thousands) for each input example and this sparsely activated neural network is shown to simultaneously achieve higher learning accuracy and lower computation than the state-of-the-art single model approach. Multi-gate MoEs are also used to  explore the relationships in multi-tasks~\cite{multiTaskRelation2018kdd}.
Multi-modal deep learning\ifmaximal~\cite{multiModal2017PAMI,oneModel2017arxiv}\fi~has also been developed as a progress towards artificial general intelligence. It incorporates features in various modality (e.g., video, voice, text) to activate partial or all neural network models to accomplish one or more tasks. In~\cite{trend2020isscc}, the combination of sparsely activated models with many tasks and modalities are projected as promising future directions for AI researches.

\section{Solving Complicated Problems with Sparsity} \label{sec:solutions_to_complicated_problems}

\begin{figure}[t]
\begin{center}
   \includegraphics[width=1.0\linewidth]{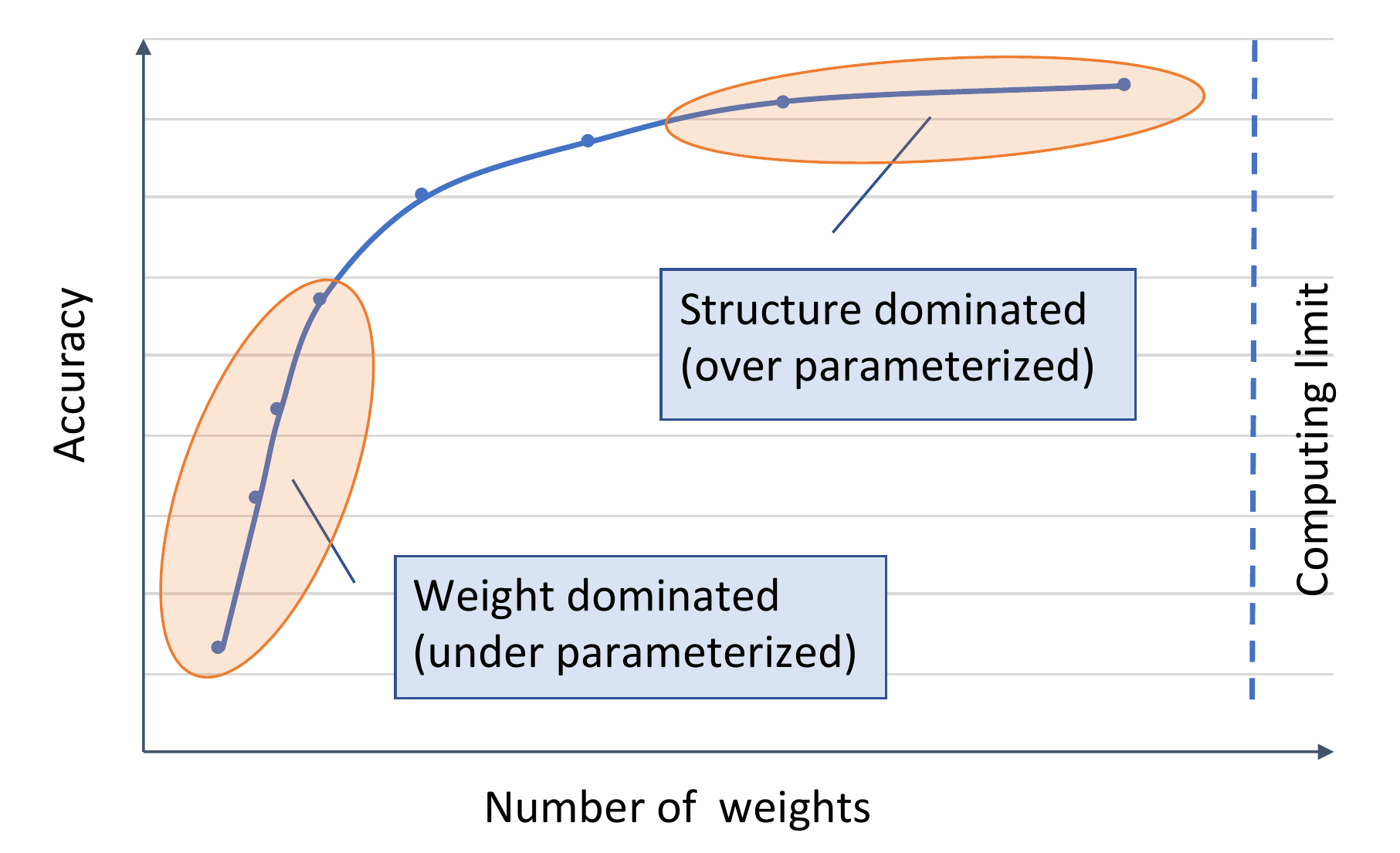}
\end{center}
   \caption{Illustration of the relations between model accuracy and the number of weights in a scalable model architecture. }
\label{fig:param_vs_accuracy}
\end{figure}

In many modern vision models, such as MobileNet family and EfficientNet, the number of weights of the models can be controlled by one scaling factor. Thus, the same model architecture can be applied to both edges (when the scaling factor is small) and servers (when the scaling factor is large). Fig.~\ref{fig:param_vs_accuracy} illustrates the relationship between the number of weights and the model accuracy in such a typical model. One interesting observation is that the accuracy of the model increases rapidly along with the number of weights in a small model, as shown on the left side of Fig.~\ref{fig:param_vs_accuracy}. Such models are usually under-parameterized, and we call this region {\em weight dominated} region, as most weights are effectively used and any change in the number of weights affects the model quality. On the other hand, towards the right side of Fig.~\ref{fig:param_vs_accuracy}, the model accuracy is insensitive to the number of weights. The model accuracy is more limited by the macro-structure rather than the number of weights. We call this region {\em structure dominated}, and models in this region are usually over-parameterized.

In most of the model designs for training and inference on servers, 
one can usually increase the number of weights to improve accuracy and obtain a model in the structure dominated region. This, however, relies on the fact that the model is within the available computing limit, {\it i.e.} the model can be trained within reasonable amount of time using the available computing resources. This is shown in the rightmost line in Fig.~\ref{fig:param_vs_accuracy}.

\begin{figure}[t]
\begin{center}
   \includegraphics[width=1.0\linewidth]{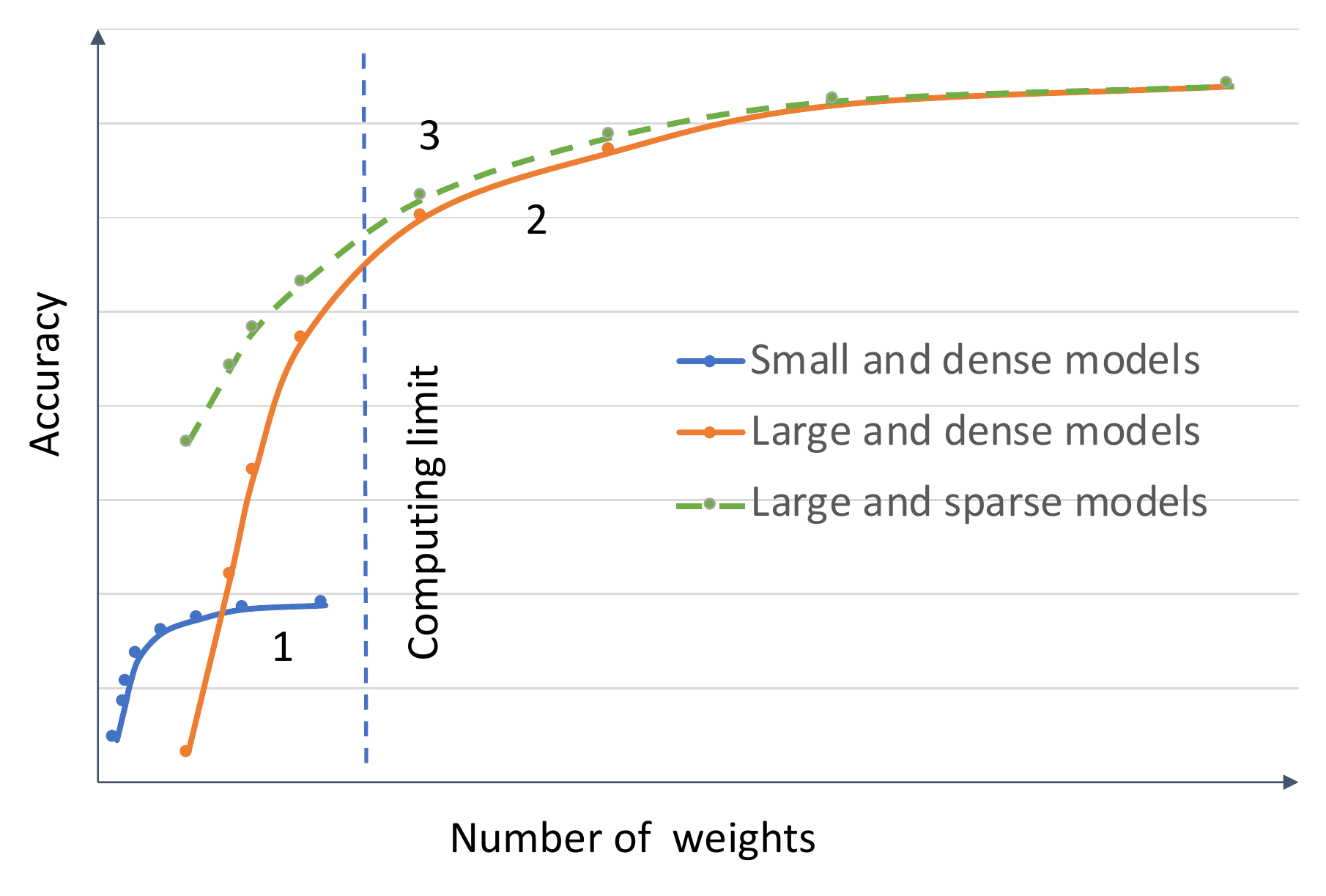}
\end{center}
   \caption{Illustration of the relations between model accuracy and the number of weights. Line 1 is a hypothetical series of dense models  within the existing computing limit. Line 2 is a hypothetical  series of large and dense models well beyond the existing computing limit. Line 3 is a hypothetical series of large and sparse models derived from the largest model in Line 2.}
\label{fig:large_param_vs_accuracy}
\end{figure}

However, this approach is insufficient to solve complicated problems. 
\ifmaximal As described in Section~\ref{sec:complicated_problems}, complicated \fi 
problems are limited by the computing resources and the quality of the results are unsatisfactory. Fig.~\ref{fig:large_param_vs_accuracy} hypothetically illustrates the relations between the accuracy and the number of weights in a neural network model for a complicated problem. Line~1 is a hypothetical series of models whose structure dominated region is below the existing computing limit. As the computing capability increases in the future, it is reasonable to assume that a much larger model with much higher accuracy can be found. It is also reasonable to assume that this series of models also contain the weight dominated region and the structure dominated region, as shown by Line 2 in Fig.~\ref{fig:large_param_vs_accuracy}. Even though the structure dominated region of the series of models is well beyond the existing computing limit, the weight dominated region may still be partially below it.
Thus, Line~2 may potentially produce higher accuracy than Line~1.
Searching for a larger model in the weight dominated region under the  computing limit may generate fruitful research results.

On the other hand, the sparse models described in Section~\ref{sec:static_sparsity} mostly focus on the structure dominated region, as the quality of the sparse model is compared to the original dense model that it is generated from, rather than a dense model with the same FLOPs or number of weights. We hypothesize that in the weight dominated region, sparse models usually result in much higher accuracy than the dense models with the same FLOPs or number of weights. In the structure dominated region, the accuracy difference is not obvious. 

\begin{figure}[t]
\begin{center}
   \includegraphics[width=1.0\linewidth]{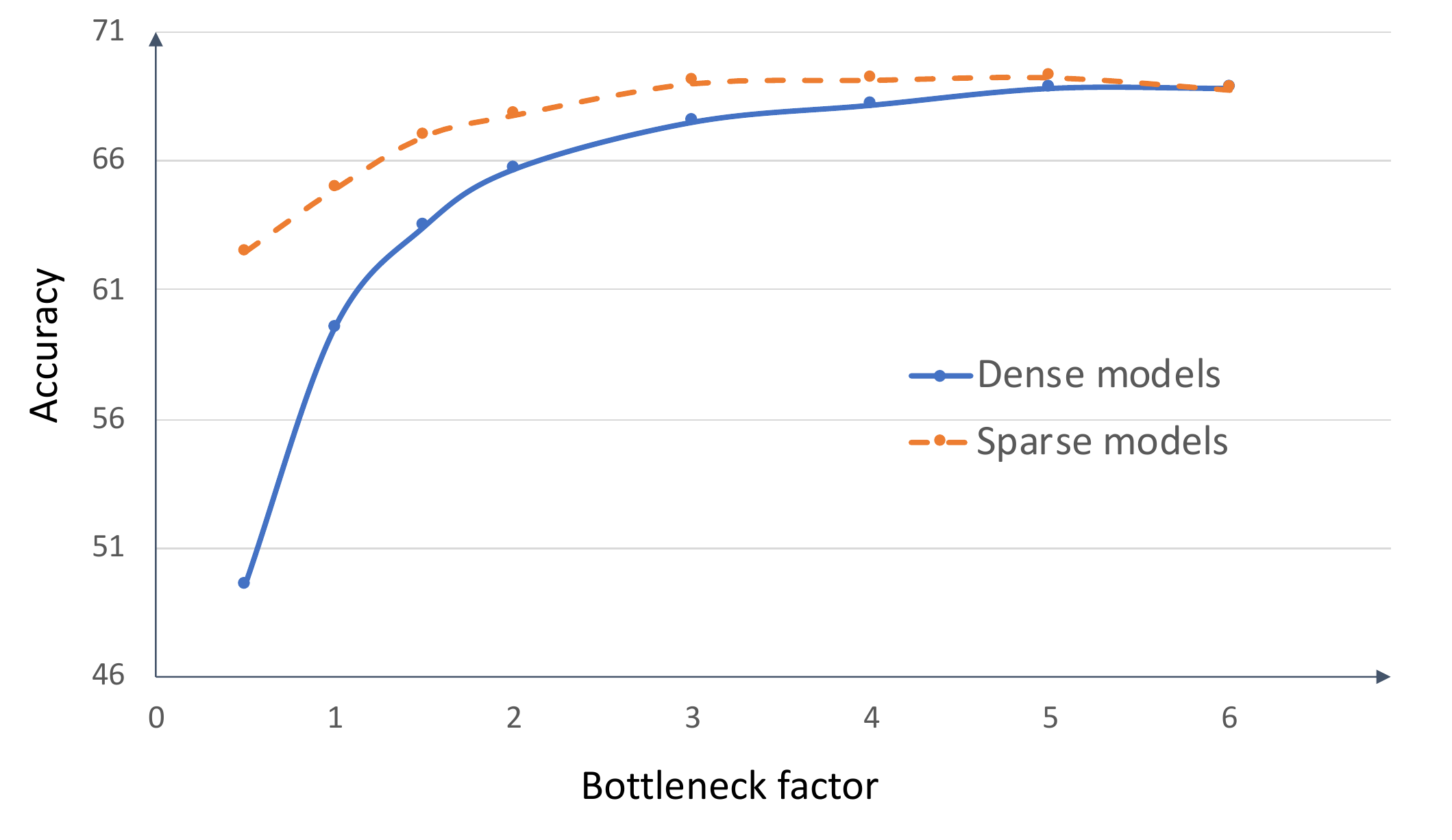}
\end{center}
   \caption{Comparison of the validation accuracy of the sparse and dense models based on MobileNetV2 with the same number of weights and FLOPs.}
\label{fig:mobilenetv2_sparsity}
\end{figure}

In order to test our hypothesis, we have modified the MobileNetV2 0.35 model~\cite{mobileNetv22018cvpr}. In a reverse bottleneck layer of MobileNetV2, the number of channels expands 6 times internally. We have created a series of models by keeping the number of the expanded channels the same, while varying the number of the bottleneck channels to be $0.5$, $1$ (the original MobileNetV2 model), $1.5$, $2$, $3$, $4$, $5$, and $6$ (the number of channels of the bottleneck layer is the same as the expanded layer). We have also generated a series of irregular sparse models with the same structure as the largest model above, but varying the sparsity so that the number of weights and  FLOPs are the same as the above dense models. We have trained the model using the ImageNet dataset and present the validation accuracy in Fig.~\ref{fig:mobilenetv2_sparsity}. 

Fig.~\ref{fig:mobilenetv2_sparsity} confirms our hypothesis that 
the accuracy difference between the sparse and dense models with the same FLOPs is larger in the weight dominated region than in the structure dominated region.
This may be due to the fact that in the dense model, the number of activation channels is reduced along with the number of weight channels in the bottleneck layer.
On the other hand, in the sparse model, although the number of FLOPS is reduced by sparsifying the weight channels, the number of channels in the activation remains the same, which leaves the sparse models more freedom to select different activation channels for different neurons. 

It is reasonable to extrapolate the same conclusion to large models beyond the existing computing limit:  the sparse models pruned from large and dense models achieve much higher accuracy in the weight dominated region than the corresponding dense models with the same number of weights or FLOPs. 

Since the research of sparsity focuses on the weight dominated region with an acceptable accuracy drop from the dense and large model, the achievable sparsity level can be much higher than the state-of-the-art described in Section~\ref{sec:static_sparsity}. 

In Fig.~\ref{fig:large_param_vs_accuracy}, Line 3 is a hypothetical series of sparse models, and Line 2 is the corresponding series of dense models. Thus, it is more beneficial to search large and sparse models directly.

However, due to the computing limit, the existing pruning mechanism described in Section~\ref{sec:static_sparsity} is no longer applicable, because the pre-trained large and dense model is well beyond the computing limit. 
Therefore, it is necessary to derive the large and sparse models by growing the small models directly, with all operations under the computing limit. However, there are only few researches in this direction~\cite{dai2019nest,du2019cgap}.

We consider this an important research area. 
The complicated problems benefit more than the simple problems by using large and sparse models to improve the accuracy.
This also indicates that searching for sparse models is not only for the purpose of improving computation efficiency, but also for the model architecture exploration. Thus, finding an optimal sparse structure of a neural network model can be framed as a neural architecture search problem.

\section{Software Frameworks on Sparse Computation} \label{sec:sparse_software}

Popular machine learning frameworks such as PyTorch~\ifmaximal\cite{paszke2019pytorch}~\fi and Tensorflow~\ifmaximal\cite{abadi2015tensorflow}~\fi have integrated sparse computation natively. 
\ifmaximal
The deep scalable sparse tensor network engine (DSSTNE)~\cite{dsstne} is a sparse ML framework designed to train and inference recommendation models with sparse inputs.
\fi
Those frameworks rely on the sparse kernel implementations to deliver fast sparse inference and training.

There have been a long history of efforts implementing efficient sparse kernels on existing computing platforms such as CPUs and GPUs. 
There are three levels of sparse kernels: level one is the sparse vector dense vector (spVV) operations; level two is the sparse matrix dense vector (spMV) operations; and level three is the sparse matrix dense matrix (spMM) operations. The sparse-sparse  operations are also popular in ultra-sparse matrices. 
The Intel math kernel library (MKL)~\ifmaximal\cite{mkl}~\fi provides efficient sparse BLAS on Intel CPUs, and cuSPARSE~\ifmaximal\cite{cusparse}~\fi is the sparse library on NVIDIA GPUs. 

\ifmaximal

The efficiency of the sparse kernels is determined by the sparse matrix representation, the software optimization, and the hardware architecture.

Sparse matrices are often represented using compact formats. As in a sparse matrix, the majority of the values are zero. Some widely used formats are compressed sparse row (CSR), compressed sparse column (CSC), block compressed row (BSR), coordinate list (COO), and list of lists (LIL). The choice of the formats is application-specific and may impact performance~\cite{filippone2017sparse}. Those compact formats require explicitly or implicitly indexing the coordinates of the non-zero values, which translates to indirect memory access and storage overhead. Comparing with the dense counterparts, those overheads fundamentally limit the performance improvement at low sparsity levels. 

Commercial sparse libraries such as MKL and cuSPARSE
\fi
 are widely used in scientific computation. It often contains extremely sparse matrices where the number of non-zero values is far less than one percent of the total number of values~\cite{grossman2016mvsurvey}, {\it i.e.} the sparsity level is much greater than 99\%.
At this level, the storage overhead of the indices and the indirect memory accesses are negligible comparing with the dense matrix computation. 

In neural networks, however, the sparsity of the matrices may be much less than 99\%. 
For example, it is desirable to achieve performance improvement when the sparsity level is around 70\%.
This brings a lot of challenges to the sparse kernel designs. The block sparse library by OpenAI~\cite{gray2017gpu} explores efficient sparse implementations on GPUs at the block granularity of 8$\times$8 or more. 
\ifmaximal
Sparsity at block level is suitable for computing sparse matrices exhibiting the characteristics of small world networks~\cite{watts1999swn}.
\fi
SparseTrain~\cite{gong2019sparsetrain} leverages the dynamic sparsity introduced by $ReLU$ and obtains speed ups on convolution operators at low sparsity levels on Intel architecture with AVX-512 extensions. Such research marginally improves the execution efficiency on existing hardware.

\section{Sparse Hardware Accelerators} \label{sec:sparse_hardware}

The state-of-the-art pruning algorithms described in Section~\ref{sec:sparse} is capable of introducing 70\% to 99\% sparsity, which translates to 3$\times$ to 100$\times$ theoretical speedup. However, the reduced number of weights and MACs may not translate proportionally to the wall clock time savings, mainly because the hardware architectures are not optimized for the sparsity at this level. 

The CPUs and GPUs are heavily optimized for the dense matrix computation. The AVX-512 module in CPU is capable of computing the same floating point (FP) operation on 16 single precision data per cycle. A Tensor Core in NVIDIA GPUs has a throughput of computing 4$\times$4 matrix multiplications per cycle. The CUDA warp is most efficient executing the same instruction on 32 data. The newer TPU uses 128$\times$128 systolic array as the matrix multiplication engine, which is most efficient performing large matrix multiplications due to its rigid structure.

However, when performing sparse matrix multiplications, those wide execution units can only utilize a small fraction of the peak performance due to  indirect memory access. Thus, many accelerators for sparse matrix multiplications have been proposed. 

\ifmaximal
To speed up the large sparse matrices used in scientific computation, OuterSPACE~\cite{pal2018outerspace} executes sparse matrix multiplication via outer products to leverage input reuse instead of the inner products often used on dense matrices.  SpArch further obtains output reuse by merging partial matrix on-chip \cite{zhang2020sparch}. These approaches, however, are more suitable for ultra-sparse matrices with sparsity level much higher than 99\%.






With a much lower achievable sparsity level in the sparse neural networks, early
\fi
sparse accelerators integrate element-wise weight sparsity through compressed storage and computing skip of zero weights~\ifmaximal\cite{han2016eie, han2017ese, zhang2016cambricon, parashar2017scnn}\fi. However, the high indexing overhead of irregular element-wise sparsity motivates the structured weight sparsity. Scalpel \cite{yu2017scalpel} proposes SIMD-aware weight sparsity, maintaining non-zero weights in aligned fixed-size groups to fully utilize the SIMD units, on low-parallelism platforms for more regular execution pattern. To take this one step further, column-wise weight sparsity \cite{kung2018packing}, block-wise weight sparsity \cite{zhou2018cambricon}, or intra-block structured weight sparsity \cite{deng2018permdnn, zhu2019sparse} are leveraged for more aggressive performance improvement. Note that the sparse weights remain static after training which simplifies the architecture design. Since the neuron activations evolve with different inputs, the dynamic activation sparsity is more difficult to exploit than the static weight sparsity.

On the dynamic sparsity side, a lot of the existing accelerators leverage the input zeros from the previous layer produced by $ReLU$ function. Eyeriss \cite{chen2017eyeriss} and SCNN \cite{parashar2017scnn} compress the zero activations for memory reduction and use computation-gating for energy saving. EIE \cite{han2016eie}, Cnvlutin \cite{Albericio2016}, NullHop \cite{aimar2018nullhop}\ifmaximal, and others \cite{ZeNA2017DandT,zhu2018sparsenn}\fi ~further skip the cycles involving zero inputs for both energy saving and execution acceleration. SparTen \cite{gondimalla2019sparten} with weight sorting and SNAP \cite{zhang2019SNAP} with associative index matching attempt to mitigate the MAC under-utilization due to load imbalance in sparse weight and input processing. 
Diffy \cite{mahmoud2018diffy} exploits input spatial similarity for computation saving and data compression.
\ifmaximal
Some accelerators also skip the negative outputs that would be rectified to zeros by $ReLU$ function. Y. Lin et al. \cite{lin2017predictivenet} and M. Song et al. \cite{song2018prediction} leverage the high-order bits data to approximate the output activations; SnaPEA \cite{akhlaghi2018snapea} proposes to reorder the sequence of MACs and early-terminate the execution upon predicting negative output. Instead of the fine-grain bit-level operations~\cite{song2018prediction,lin2017predictivenet} and the early calculation of predictable outputs \cite{akhlaghi2018snapea}, dynamic channel gating \cite{hua2019boosting} can also be used to reduce computations. 
\fi

Most of the above-mentioned accelerators design for improving the execution efficiency on inference workload. A lot of them target edge devices with fewer processing elements (PEs). 

As described in Section~\ref{sec:solutions_to_complicated_problems}, when targeting complicated problems with sparse models close to the compute limit, it is preferable to grow from a smaller network. In this process, it is important to be able to train a sparse network directly, so that the entire training can be done below the computing limit. Thus, an efficient sparse training hardware is essential to solve the complicated problems. 

A few recent works have started to target sparse training on the server side. SIGMA~\cite{qin2020sigma} proposes flexible routing and reduction networks in hardware to reduce the indirect memory reference overhead. Its 128$\times$128 Flex-DPE matches the size of a TPU but delivers higher utilization of the PEs. We consider this an important area of research and more attention from the research community is warranted. 
\section{Future Research Directions} \label{sec:research_directions}

The current research on sparsity has been primarily focused on the structure dominated models targeting simple problems. The efforts on the sparse computation frameworks and the sparse hardware accelerators mainly try to improve the computation efficiency, {\it i.e.} reducing the amount of computation from a large and dense model. 

In order to effectively and efficiently solve the complicated problems, we need to consider sparsity as a first class generic model exploration methodology, which is necessary due to the following reasons:
\begin{itemize}
    \item The large and sparse models close to the computing limit can only be grown from the smaller models, rather than being pruned from the even larger models. 
    \item The sparse algorithms may be universally applied to different application domains, such as vision, NLP, and recommendation, and it may achieve acceptable results. 
    \item A sparse model structure may be specialized on the training dataset. Thus, exploration is needed to update the sparse model structure when transferring the model to a different application or domain. 
\end{itemize}

Since the existing CPU and GPU implementations may not fulfill the need of the training and inference the sparse models, new hardware accelerators and the corresponding software frameworks need to be developed to speed up the run time. 
Working with the models for complicated problems, the iteration speed is critical. It is essential to design new sparse training accelerators specifically targeting the sparsity level those models likely falling into ({\it e.g., }90-99\%). 

To make significant progress on sparse model researches, we need to make coordinated breakthroughs in algorithm, software, and hardware. These three disciplines are tightly intertwined, and a co-design approach is preferred. 

\ifmaximal
\section{conclusion} \label{sec:conclusion}

The amount of computation we currently have limits our ability to explore some complicated problems, such as video, point cloud, transformer based NLP, and graph. 
Even though the amount of computation used to train the largest models doubles every 3.4 months, the trend is unlikely to sustain economically. 
Thus, we have expressed our view in alleviating the computation scarcity problem by focusing the future neural network designs based on the sparse matrices rather than the dense matrices. 
We have shown that in the weight dominated region, the sparse models achieve much higher accuracy than the dense models with the same number of FLOPs or weights, and we project that we may observe the same on the complicated problems. 
We need to make breakthroughs in researching the computation on sparse neural networks across the stack: from algorithm, software, to hardware, among which a new sparse training hardware is essential to facilitate fast iteration of the sparse algorithms. 
\fi

\bibliographystyle{IEEEtranS}

\begin{thebibliography}{100}
\providecommand{\url}[1]{#1}
\csname url@samestyle\endcsname
\providecommand{\newblock}{\relax}
\providecommand{\bibinfo}[2]{#2}
\providecommand{\BIBentrySTDinterwordspacing}{\spaceskip=0pt\relax}
\providecommand{\BIBentryALTinterwordstretchfactor}{4}
\providecommand{\BIBentryALTinterwordspacing}{\spaceskip=\fontdimen2\font plus
\BIBentryALTinterwordstretchfactor\fontdimen3\font minus
  \fontdimen4\font\relax}
\providecommand{\BIBforeignlanguage}[2]{{%
\expandafter\ifx\csname l@#1\endcsname\relax
\typeout{** WARNING: IEEEtranS.bst: No hyphenation pattern has been}%
\typeout{** loaded for the language `#1'. Using the pattern for}%
\typeout{** the default language instead.}%
\else
\language=\csname l@#1\endcsname
\fi
#2}}
\providecommand{\BIBdecl}{\relax}
\BIBdecl

\bibitem{abadi2015tensorflow}
M.~Abadi, P.~Barham, J.~Chen \emph{et~al.}, ``{TensorFlow}: A system for
  large-scale machine learning,'' in \emph{Proc. Symp. Operating Systems Design
  \& Implementation}, 2016, pp. 265--283.

\bibitem{AcceleratingCN2018VLSI}
T.~Abtahi, C.~Shea, A.~M. Kulkarni, and T.~Mohsenin, ``Accelerating
  convolutional neural network with {FFT} on embedded hardware,'' \emph{IEEE
  Trans. VLSI Systems}, vol.~26, no.~9, pp. 1737--1749, Sept. 2018.

\bibitem{aimar2018nullhop}
A.~Aimar, H.~Mostafa, E.~Calabrese \emph{et~al.}, ``Nullhop: A flexible
  convolutional neural network accelerator based on sparse representations of
  feature maps,'' \emph{IEEE Trans. Neural Networks and Learning Systems},
  no.~99, pp. 1--13, 2018.

\bibitem{akhlaghi2018snapea}
V.~Akhlaghi, A.~Yazdanbakhsh, K.~Samadi, R.~K. Gupta, and H.~Esmaeilzadeh,
  ``Snapea: Predictive early activation for reducing computation in deep
  convolutional neural networks,'' in \emph{Proc. Int. Symp. Computer
  Architecture}, 2018, pp. 662--673.

\bibitem{Albericio2016}
J.~Albericio, P.~Judd, T.~Hetherington, T.~Aamodt, N.~E. Jerger, and
  A.~Moshovos, ``Cnvlutin: Ineffectual-neuron-free deep neural network
  computing,'' in \emph{Proc. Int. Symp. Computer Architecture}, 2016, pp.
  1--13.

\bibitem{dsstne}
\BIBentryALTinterwordspacing
Amazon, ``Amazon {DSSTNE}: deep scalable sparse tensor network engine.''
  [Online]. Available: \url{https://github.com/amzn/amazon-dsstne}
\BIBentrySTDinterwordspacing

\bibitem{ActvationPrune2017GlobalSIP}
A.~Ardakani, C.~Condo, and W.~J. Gross, ``Activation pruning of deep
  convolutional neural networks,'' in \emph{IEEE Global Conf. Signal and
  Information Processing (GlobalSIP)}, 2017, pp. 1325--1329.

\bibitem{multiModal2017PAMI}
T.~Baltrusaitis, C.~Ahuja, and L.-P. Morency, ``Multimodal machine learning: a
  survey and taxonomy,'' \emph{IEEE Trans. Pattern Analysis and Machine
  Intelligence}, vol.~41, no.~2, p. 423–443, Feb. 2019.

\bibitem{SacrificingAF2018icann}
K.~Berestizshevsky and G.~Even, ``Sacrificing accuracy for reduced computation:
  Cascaded inference based on softmax confidence,'' \emph{arXiv preprint
  arXiv:1805.10982}, 2018.

\bibitem{AdaptiveNN2017icml}
T.~Bolukbasi, J.~Wang, O.~Dekel, and V.~Saligrama, ``Adaptive neural networks
  for efficient inference,'' in \emph{Proc. Int. Conf. Machine Learning}, 2017,
  pp. 527--536.

\bibitem{onceForAll2020iclr}
H.~Cai, C.~Gan, and S.~Han, ``Once for all: Train one network and specialize it
  for efficient deployment,'' in \emph{Proc. Int. Conf. Learning
  Representations}, 2020.

\bibitem{campos2018skip}
V.~Campos, B.~Jou, X.~G. i~Nieto, J.~Torres, and S.-F. Chang, ``Skip {RNN}:
  Learning to skip state updates in recurrent neural networks,'' in \emph{Proc.
  Int. Conf. Learning Representations}, 2018.

\bibitem{cao2019seernet}
S.~Cao, L.~Ma, W.~Xiao, C.~Zhang, Y.~Liu, L.~Zhang, L.~Nie, and Z.~Yang,
  ``Seer{N}et: Predicting convolutional neural network feature-map sparsity
  through low-bit quantization,'' in \emph{Proc. Conf. Computer Vision and
  Pattern Recognition}, 2019, pp. 11\,216--11\,225.

\bibitem{Carroll1970AnalysisOI}
J.~D. Carroll and J.-J. Chang, ``Analysis of individual differences in
  multidimensional scaling via an n-way generalization of “eckart-young”
  decomposition,'' \emph{Psychometrika}, vol.~35, pp. 283--319, 1970.

\bibitem{candelinc1980}
J.~D. Carroll, S.~Pruzansky, and J.~B. Kruskal, ``Candelinc: A general approach
  to multidimensional analysis of many-way arrays with linear constraints on
  parameters,'' \emph{Psychometrika}, vol.~45, pp. 3--24, 1980.

\bibitem{chen2017eyeriss}
Y.-H. Chen, T.~Krishna, J.~S. Emer, and V.~Sze, ``Eyeriss: An energy-efficient
  reconfigurable accelerator for deep convolutional neural networks,''
  \emph{IEEE J. Solid-State Circuits}, vol.~52, no.~1, pp. 127--138, 2017.

\bibitem{choy20194d}
C.~Choy, J.~Gwak, and S.~Savarese, ``{4D} spatio-temporal convnets: Minkowski
  convolutional neural networks,'' in \emph{Proc. Conf. Computer Vision and
  Pattern Recognition}, 2019, pp. 3075--3084.

\bibitem{dai2019nest}
X.~Dai, H.~Yin, and N.~K. Jha, ``Nest: A neural network synthesis tool based on
  a grow-and-prune paradigm,'' in \emph{IEEE Trans. Computers}, vol.~68,
  no.~10, Oct. 2019, pp. 1487--1497.

\bibitem{trend2020isscc}
J.~Dean, ``The deep learning revolution and its implications for computer
  architecture and chip design,'' \emph{arXiv preprint arXiv:1911.05289}, 2019.

\bibitem{degenne1999introducing}
A.~Degenne and M.~Fors{\'e}, \emph{Introducing social networks}.\hskip 1em plus
  0.5em minus 0.4em\relax Sage, 1999.

\bibitem{deng2018permdnn}
C.~Deng, S.~Liao, Y.~Xie, K.~K. Parhi, X.~Qian, and B.~Yuan, ``{PermDNN}:
  Efficient compressed {DNN} architecture with permuted diagonal matrices,'' in
  \emph{Proc. Int. Symp. Microarchitecture}, 2018, pp. 189--202.

\bibitem{dong2017complicated}
X.~Dong, J.~Huang, Y.~Yang, and S.~Yan, ``More is less: A more complicated
  network with less inference complexity,'' \emph{arXiv preprint
  arXiv:1703.08651}, 2017.

\bibitem{du2019cgap}
X.~Du, Z.~Li, and Y.~Cao, ``{CGaP}: Continuous growth and pruning for efficient
  deep learning,'' \emph{arXiv preprint arXiv:1905.11533}, 2019.

\bibitem{SpatiallyAC2017cvpr}
M.~Figurnov, M.~D. Collins, Y.~Zhu, L.~Zhang, J.~Huang, D.~P. Vetrov, and
  R.~Salakhutdinov, ``Spatially adaptive computation time for residual
  networks,'' \emph{Proc. Conf. Computer Vision and Pattern Recognition}, pp.
  1790--1799, 2017.

\bibitem{filippone2017sparse}
S.~Filippone, V.~Cardellini, D.~Barbieri, and A.~Fanfarillo, ``Sparse
  matrix-vector multiplication on {GPGPUs},'' \emph{ACM trans. Mathematical
  Software}, vol.~43, no.~4, pp. 1--49, Mar. 2017.

\bibitem{lotteryTicket2018iclr}
J.~Frankle and M.~Carbin, ``The lottery ticket hypothesis: Training pruned
  neural networks,'' in \emph{Proc. Int. Conf. Learning Representations}, 2019.

\bibitem{gao2018dynamic}
X.~Gao, Y.~Zhao, Łukasz Dudziak, R.~Mullins, and C.~zhong Xu, ``Dynamic
  channel pruning: Feature boosting and suppression,'' \emph{arXiv preprint
  arXiv:1810.05331}, 2018.

\bibitem{gondimalla2019sparten}
A.~Gondimalla, N.~Chesnut, M.~Thottethodi, and T.~N. Vijaykumar, ``Sparten: A
  sparse tensor accelerator for convolutional neural networks,'' in \emph{Proc.
  Int. Symp. Microarchitecture}, 2019, p. 151–165.

\bibitem{gong2019sparsetrain}
Z.~Gong, H.~Ji, C.~Fletcher, C.~Hughes, and J.~Torrellas, ``{SparseTrain}:
  Leveraging dynamic sparsity in training {DNNs} on general-purpose {SIMD}
  processors,'' \emph{arXiv preprint arXiv:1911.10175}, 2019.

\bibitem{gray2017gpu}
S.~Gray, A.~Radford, and D.~P. Kingma, ``{GPU} kernels for block-sparse
  weights,'' Technical report, OpenAI, Tech. Rep., 2017.

\bibitem{grossman2016mvsurvey}
M.~Grossman, C.~Thiele, M.~Araya-Polo, F.~Frank, F.~O. Alpak, and V.~Sarkar,
  ``A survey of sparse matrix-vector multiplication performance on large
  matrices,'' \emph{arXiv preprint arXiv:1608.00636}, 2016.

\bibitem{han2017ese}
S.~Han, J.~Kang, H.~Mao, Y.~Hu, X.~Li, Y.~Li, D.~Xie, H.~Luo, S.~Yao, Y.~Wang,
  H.~Yang, and W.~J. Dally, ``{ESE}: Efficient speech recognition engine with
  sparse {LSTM} on {FPGA},'' in \emph{Proc. Int. Symp. Field-Programmable Gate
  Arrays}, 2017, pp. 75--84.

\bibitem{han2016eie}
S.~Han, X.~Liu, H.~Mao, J.~Pu, A.~Pedram, M.~A. Horowitz, and W.~J. Dally,
  ``{EIE}: efficient inference engine on compressed deep neural network,'' in
  \emph{Proc. Int. Symp. Computer Architecture}, 2016, pp. 243--254.

\bibitem{Han2016a}
S.~Han, H.~Mao, and W.~J. Dally, ``Deep compression: Compressing deep neural
  networks with pruning, trained quantization and huffman coding,'' in
  \emph{Proc. Int. Conf. Learning Representations}, 2016.

\bibitem{han2015learning}
S.~Han, J.~Pool, J.~Tran, and W.~Dally, ``Learning both weights and connections
  for efficient neural network,'' in \emph{Proc. Conf. Neural Information
  Processing Systems}, 2015, pp. 1135--1143.

\bibitem{parafac1972ucla}
R.~A. Harshman, ``{PARAFAC2}: Mathematical and technical notes,'' \emph{UCLA
  Working Papers in Phonetics}, vol.~22, pp. 30--44, 1972b.

\bibitem{Harshman96uniquenessproof}
R.~A. Harshman, Margaret, and E.~Lundy, ``Uniqueness proof for a family of
  models sharing features of tucker’s three-mode factor analysis and
  parafac/candecomp,'' \emph{Psychometrika}, 1996.

\bibitem{he2015rectifiers}
K.~He, X.~Zhang, S.~Ren, and J.~Sun, ``Delving deep into rectifiers: surpassing
  human-level performance on {I}mager{N}et classification,'' in \emph{Proc.
  Int. Conf. Computer Vision}, 2015, pp. 1026--1034.

\bibitem{Polyadic1927JMP}
F.~L. Hitchcock, ``The expression of a tensor or a polyadic as a sum of
  products,'' \emph{J. of Mathematics and Physics}, vol.~6, no. 1-4, pp.
  164--189, 1927.

\bibitem{Pwaytensor1928JMP}
{Hitchcock, Frank L.}, ``Multiple invariants and generalized rank of a p-way
  matrix or tensor,'' \emph{J. of Mathematics and Physics}, vol.~7, no. 1-4,
  pp. 39--79, 1928.

\bibitem{hua2019boosting}
W.~Hua, Y.~Zhou, C.~De~Sa, Z.~Zhang, and G.~E. Suh, ``Boosting the performance
  of {CNN} accelerators with dynamic fine-grained channel gating,'' in
  \emph{Proc. Int. Symp. Microarchitecture}, 2019, p. 139–150.

\bibitem{hua2019channel}
W.~Hua, Y.~Zhou, C.~M. De~Sa, Z.~Zhang, and G.~E. Suh, ``Channel gating neural
  networks,'' in \emph{Proc. Conf. Neural Information Processing Systems},
  2019, pp. 1884--1894.

\bibitem{MultiScaleDN2017iclr}
G.~Huang, D.~Chen, T.~Li, F.~Wu, L.~van~der Maaten, and K.~Q. Weinberger,
  ``Multi-scale dense networks for resource efficient image classification,''
  in \emph{Proc. Int. Conf. Learning Representations}, 2017.

\bibitem{quantizeDNN2017MLR}
I.~Hubara, M.~Courbariaux, D.~Soudry, R.~El-Yaniv, and Y.~Bengio, ``Quantized
  neural networks: Training neural networks with low precision weights and
  activations,'' \emph{J. Machine Learning Research}, vol.~18, no.~1, p.
  6869–6898, Jan. 2017.

\bibitem{mkl}
\BIBentryALTinterwordspacing
Intel, ``Intel math kernel library.'' [Online]. Available:
  \url{https://software.intel.com/en-us/mkl}
\BIBentrySTDinterwordspacing

\bibitem{Gambel2017iclr}
E.~Jang, S.~Gu, and B.~Poole, ``Categorical reparameterization with
  gumbel-softmax,'' \emph{arXiv preprint arXiv:1611.01144}, 2017.

\bibitem{oneModel2017arxiv}
L.~Kaiser, A.~N. Gomez, N.~Shazeer, A.~Vaswani, N.~Parmar, L.~Jones, and
  J.~Uszkoreit, ``One model to learn them all,'' \emph{arXiv preprint
  arXiv:1706.05137}, 2017.

\bibitem{shallowdeepnetworks2019icml}
Y.~Kaya, S.~Hong, and T.~Dumitras, ``Shallow-deep networks: Understanding and
  mitigating network overthinking,'' in \emph{Proc. Int. Conf. Machine
  Learning}, Jun 2019.

\bibitem{ZeNA2017DandT}
D.~Kim, J.~Ahn, and S.~Yoo, ``{ZeNA}: Zero-aware neural network accelerator,''
  \emph{IEEE Design \& Test}, vol.~35, no.~1, pp. 39--46, 2018.

\bibitem{FPGAzeroSkipping2018ISRSP}
D.~Kim, S.~Kim, and S.~Yoo, ``{FPGA} prototyping of low-precision zero-skipping
  accelerator for neural networks,'' in \emph{Int. Symp. Rapid System
  Prototyping (RSP)}, 2018, pp. 104--110.

\bibitem{Krizhevsky2012}
A.~Krizhevsky, I.~Sutskever, and G.~E. Hinton, ``{ImageNet} classification with
  deep convolutional neural networks,'' in \emph{Proc. Conf. Neural Information
  Processing Systems}, Dec. 2012, pp. 1097--1105.

\bibitem{kung2018packing}
H.~Kung, B.~McDanel, and S.~Q. Zhang, ``Packing sparse convolutional neural
  networks for efficient systolic array implementations: Column combining under
  joint optimization,'' \emph{arXiv preprint arXiv:1811.04770}, 2018.

\bibitem{winofft2016cvpr}
A.~Lavin and S.~Gray, ``Fast algorithms for convolutional neural networks,'' in
  \emph{Proc. Conf. Computer Vision and Pattern Recognition}, June 2016, pp.
  4013--4021.

\bibitem{lin2017predictivenet}
Y.~Lin, C.~Sakr, Y.~Kim, and N.~Shanbhag, ``Predictive{N}et: an
  energy-efficient convolutional neural network via zero prediction,'' in
  \emph{Proc. Int. Symp. Circuits \& Systems}, 2017, pp. 1--4.

\bibitem{liu2018dynamic}
L.~Liu, L.~Deng, X.~Hu, M.~Zhu, G.~Li, Y.~Ding, and Y.~Xie, ``Dynamic sparse
  graph for efficient deep learning,'' in \emph{Proc. Int. Conf. Learning
  Representations}, 2019.

\bibitem{Rethinking2019ICLR}
Z.~Liu, M.~Sun, T.~Zhou, G.~Huang, and T.~Darrell, ``Rethinking the value of
  network pruning,'' in \emph{Proc. Int. Conf. Learning Representations}, 2018.

\bibitem{LearningL02017iclr}
C.~Louizos, M.~Welling, and D.~P. Kingma, ``Learning sparse neural networks
  through l0 regularization,'' in \emph{Proc. Int. Conf. Learning
  Representations}, 2018.

\bibitem{multiTaskRelation2018kdd}
J.~Ma, Z.~Zhao, X.~Yi, J.~Chen, L.~Hong, and E.~H. Chi, ``Modeling task
  relationships in multi-task learning with multi-gate mixture-of-experts,'' in
  \emph{Proc. Int. Conf. Knowledge Discovery \& Data Mining}, 2018, p.
  1930–1939.

\bibitem{ma2019pconv}
X.~Ma, F.-M. Guo, W.~Niu, X.~Lin, J.~Tang, K.~Ma, B.~Ren, and Y.~Wang,
  ``{PCONV}: The missing but desirable sparsity in {DNN} weight pruning for
  real-time execution on mobile devices,'' \emph{arXiv preprint
  arXiv:1909.05073}, 2019.

\bibitem{mahmoud2018diffy}
M.~Mahmoud, K.~Siu, and A.~Moshovos, ``Diffy: a d{\'e}j{\`a} vu-free
  differential deep neural network accelerator,'' in \emph{Proc. Int. Symp.
  Microarchitecture}, 2018, pp. 134--147.

\bibitem{makantasis2015deep}
K.~Makantasis, K.~Karantzalos, A.~Doulamis, and N.~Doulamis, ``Deep supervised
  learning for hyperspectral data classification through convolutional neural
  networks,'' in \emph{Int. Geoscience and Remote Sensing Symp.}, 2015, pp.
  4959--4962.

\bibitem{mikami2018sgd}
H.~Mikami, H.~Suganuma, P.~U-chupala, Y.~Tanaka, and Y.~Kageyama, ``Massively
  distributed {SGD}: imagenet/resnet-50 training in a flash,'' \emph{arXiv
  preprint arXiv:1811.05233}, 2018.

\bibitem{mnih2013playing}
V.~Mnih, K.~Kavukcuoglu, D.~Silver, A.~Graves, I.~Antonoglou, D.~Wierstra, and
  M.~Riedmiller, ``Playing atari with deep reinforcement learning,''
  \emph{arXiv preprint arXiv:1312.5602}, 2013.

\bibitem{neil2017delta}
D.~Neil, J.~H. Lee, T.~Delbruck, and S.-C. Liu, ``Delta networks for optimized
  recurrent network computation,'' in \emph{Proc. Int. Conf. Machine Learning},
  2017, pp. 2584--2593.

\bibitem{cusparse}
\BIBentryALTinterwordspacing
NVIDIA, ``{NVIDIA} {CUDA} sparse matrix library.'' [Online]. Available:
  \url{https://developer.nvidia.com/cusparse}
\BIBentrySTDinterwordspacing

\bibitem{aiandcompute}
\BIBentryALTinterwordspacing
{OpenAI}, ``{AI} and compute.'' [Online]. Available:
  \url{https://openai.com/blog/ai-and-compute/}
\BIBentrySTDinterwordspacing

\bibitem{trainingcost}
\BIBentryALTinterwordspacing
OpenAI, ``The staggering cost of training {SOTA} {AI} models.'' [Online].
  Available:
  \url{https://medium.com/syncedreview/the-staggering-cost-of-training-sota-ai-models-e329e80fa82}
\BIBentrySTDinterwordspacing

\bibitem{pal2018outerspace}
S.~Pal, J.~Beaumont, D.-H. Park \emph{et~al.}, ``{OuterSPACE}: an outer product
  based sparse matrix multiplication accelerator,'' in \emph{Proc. Int. Symp.
  High Performance Computer Architecture}, 2018, pp. 724--736.

\bibitem{parashar2017scnn}
A.~Parashar, M.~Rhu, A.~Mukkara, A.~Puglielli, R.~Venkatesan, B.~Khailany,
  J.~Emer, S.~W. Keckler, and W.~J. Dally, ``{SCNN}: An accelerator for
  compressed-sparse convolutional neural networks,'' in \emph{Proc. Int. Symp.
  Computer Architecture}, 2017, pp. 27--40.

\bibitem{EEquantize2018isca}
E.~Park, D.~Kim, and S.~Yoo, ``Energy-efficient neural network accelerator
  based on outlier-aware low-precision computation,'' in \emph{Proc. Int. Symp.
  Computer Architecture}, 2018, p. 688–698.

\bibitem{paszke2019pytorch}
A.~Paszke, S.~Gross, F.~Massa \emph{et~al.}, ``{PyTorch}: An imperative style,
  high-performance deep learning library,'' in \emph{Proc. Conf. Neural
  Information Processing Systems}, 2019, pp. 8026--8037.

\bibitem{FCNN2017ecml}
H.~Pratt, B.~Williams, F.~Coenen, and Y.~Zheng, ``{FCNN}: Fourier convolutional
  neural networks,'' in \emph{Machine Learning and Knowledge Discovery in
  Databases}, 2017, pp. 786--798.

\bibitem{qi2017pointnet++}
C.~R. Qi, L.~Yi, H.~Su, and L.~J. Guibas, ``Pointnet++: Deep hierarchical
  feature learning on point sets in a metric space,'' in \emph{Proc. Conf.
  Neural Information Processing Systems}, 2017, pp. 5099--5108.

\bibitem{qin2020sigma}
E.~Qin, A.~Samajdar, H.~Kwon, V.~Nadella, S.~Srinivasan, D.~Das, B.~Kaul, and
  T.~Krishna, ``{SIGMA}: A sparse and irregular {GEMM} accelerator with
  flexible interconnects for {DNN} training,'' in \emph{Proc. Int. Symp. High
  Performance Computer Architecture}, 2020.

\bibitem{radford2019language}
A.~Radford, J.~Wu, R.~Child, D.~Luan, D.~Amodei, and I.~Sutskever, ``Language
  models are unsupervised multitask learners,'' \emph{OpenAI Blog}, p.~9, 2019.

\bibitem{ren2019admm}
A.~Ren, T.~Zhang, S.~Ye, J.~Li, W.~Xu, X.~Qian, X.~Lin, and Y.~Wang,
  ``{ADMM-NN}: An algorithm-hardware co-design framework of {DNNs} using
  alternating direction methods of multipliers,'' in \emph{Proc. Int. Conf.
  Architectural Support for Programming Languages and Operating Systems}, 2019,
  pp. 925--938.

\bibitem{mobileNetv22018cvpr}
M.~Sandler, A.~Howard, M.~Zhu, A.~Zhmoginov, and L.~Chen, ``Mobile{N}et{V}2:
  Inverted residuals and linear bottlenecks,'' in \emph{Proc. Conf. Computer
  Vision and Pattern Recognition}, 2018, pp. 4510--4520.

\bibitem{scarselli2008graph}
F.~Scarselli, M.~Gori, A.~C. Tsoi, M.~Hagenbuchner, and G.~Monfardini, ``The
  graph neural network model,'' \emph{IEEE Trans. Neural Networks}, vol.~20,
  no.~1, pp. 61--80, 2008.

\bibitem{sparseMoE2017iclr}
N.~Shazeer, A.~Mirhoseini, K.~Maziarz, A.~Davis, Q.~Le, G.~Hinton, and J.~Dean,
  ``Outrageously large neural networks: The sparsely-gated mixture-of-experts
  layer,'' in \emph{Proc. Int. Conf. Learning Representations}, 2017.

\bibitem{song2018prediction}
M.~Song, J.~Zhao, Y.~Hu, J.~Zhang, and T.~Li, ``Prediction based execution on
  deep neural networks,'' in \emph{Proc. Int. Symp. Computer
  Architecture}.\hskip 1em plus 0.5em minus 0.4em\relax IEEE, 2018, pp.
  752--763.

\bibitem{summala1988risk}
H.~Summala, ``Risk control is not risk adjustment: The zero-risk theory of
  driver behaviour and its implications,'' \emph{Ergonomics}, pp. 491--506,
  1988.

\bibitem{Teer16icpr}
S.~Teerapittayanon, B.~McDanel, and H.-T. Kung, ``Branchy{N}et: Fast inference
  via early exiting from deep neural networks,'' in \emph{Proc. Int. Conf.
  Pattern Recognition}, 2016, pp. 2464--2469.

\bibitem{Mullapudi2018CVPR}
R.~Teja~Mullapudi, W.~R. Mark, N.~Shazeer, and K.~Fatahalian, ``{HydraNets}:
  Specialized dynamic architectures for efficient inference,'' in \emph{Proc.
  Conf. Computer Vision and Pattern Recognition}, 2018, pp. 8080--8089.

\bibitem{CSR1967IEEE}
W.~F. Tinney and J.~W. Walker, ``Direct solutions of sparse network equations
  by optimally ordered triangular factorization,'' \emph{Proc. IEEE}, vol.~55,
  no.~11, pp. 1801--1809, 1967.

\bibitem{Tuck1963a}
L.~R. Tucker, ``{I}mplications of factor analysis of three-way matrices for
  measurement of change,'' in \emph{{P}roblems in measuring change.}, C.~W.
  Harris, Ed.\hskip 1em plus 0.5em minus 0.4em\relax University of Wisconsin
  Press, 1963, pp. 122--137.

\bibitem{CNNadaptiveGraph2018eccv}
A.~Veit and S.~Belongie, ``Convolutional networks with adaptive inference
  graphs,'' in \emph{Proc. European Conf. Computer Vision.}, 2018, pp. 3--18.

\bibitem{SkipNet2018ECCV}
X.~Wang, F.~Yu, Z.-Y. Dou, T.~Darrell, and J.~E. Gonzalez, ``Skip{N}et:
  Learning dynamic routing in convolutional networks,'' in \emph{Proc. European
  Conf. Computer Vision.}, 2018, pp. 409--424.

\bibitem{watts1999swn}
D.~J. Watts, \emph{Small Worlds: The Dynamics of Networks between Order and
  Randomness}.\hskip 1em plus 0.5em minus 0.4em\relax Princeton University
  Press, 2004, vol.~9.

\bibitem{wen2016learning}
W.~Wen, C.~Wu, Y.~Wang, Y.~Chen, and H.~Li, ``Learning structured sparsity in
  deep neural networks,'' in \emph{Proc. Conf. Neural Information Processing
  Systems}, 2016, pp. 2074--2082.

\bibitem{winograd1980IAM}
S.~Winograd, \emph{Arithmetic Complexity of Computations}.\hskip 1em plus 0.5em
  minus 0.4em\relax Siam, 1980, vol.~33.

\bibitem{blockdrop2018cvpr}
Z.~Wu, T.~Nagarajan, A.~Kumar, S.~Rennie, L.~S. Davis, K.~Grauman, and
  R.~Feris, ``Blockdrop: Dynamic inference paths in residual networks,'' in
  \emph{Proc. Conf. Computer Vision and Pattern Recognition}, 2018.

\bibitem{yu2017scalpel}
J.~Yu, A.~Lukefahr, D.~Palframan, G.~Dasika, R.~Das, and S.~Mahlke, ``Scalpel:
  Customizing {DNN} pruning to the underlying hardware parallelism,'' in
  \emph{Proc. Int. Symp. Computer Architecture}, 2017, pp. 548--560.

\bibitem{zhang2019SNAP}
J.-F. Zhang, C.-E. Lee, C.~Liu, Y.~S. Shao, S.~W. Keckler, and Z.~Zhang,
  ``{SNAP}: A 1.67—21.55 {TOPS/W} sparse neural acceleration processor for
  unstructured sparse deep neural network inference in 16nm {CMOS},'' in
  \emph{Symp. VLSI Circuits}, 2019, pp. C306--C307.

\bibitem{zhang2016cambricon}
S.~Zhang, Z.~Du, L.~Zhang, H.~Lan, S.~Liu, L.~Li, Q.~Guo, T.~Chen, and Y.~Chen,
  ``{Cambricon-X}: An accelerator for sparse neural networks,'' in \emph{Proc.
  Int. Symp. Microarchitecture}, 2016, pp. 1--12.

\bibitem{zhang2018systematic}
T.~Zhang, S.~Ye, K.~Zhang, J.~Tang, W.~Wen, M.~Fardad, and Y.~Wang, ``A
  systematic {DNN} weight pruning framework using alternating direction method
  of multipliers,'' in \emph{Proc. European Conf. Computer Vision.}, 2018, pp.
  184--199.

\bibitem{zhang2018towards}
X.~{Zhang}, C.~{Xie}, J.~{Wang}, W.~{Zhang}, and X.~{Fu}, ``Towards memory
  friendly long-short term memory networks ({LSTMs}) on mobile {GPUs},'' in
  \emph{Proc. Int. Symp. Microarchitecture}, Oct 2018, pp. 162--174.

\bibitem{zhang2020sparch}
Z.~Zhang, H.~Wang, S.~Han, and W.~J. Dally, ``{SpArch}: Efficient architecture
  for sparse matrix multiplication,'' \emph{arXiv preprint arXiv:2002.08947},
  2020.

\bibitem{zhou2018cambricon}
X.~Zhou, Z.~Du, Q.~Guo, S.~Liu, C.~Liu, C.~Wang, X.~Zhou, L.~Li, T.~Chen, and
  Y.~Chen, ``{Cambricon-S}: Addressing irregularity in sparse neural networks
  through a cooperative software/hardware approach,'' in \emph{Proc. Int. Symp.
  Microarchitecture}, 2018, pp. 15--28.

\bibitem{zhu2018sparsenn}
J.~Zhu, J.~Jiang, X.~Chen, and C.-Y. Tsui, ``{SparseNN}: An energy-efficient
  neural network accelerator exploiting input and output sparsity,'' in
  \emph{Proc. Design Automation \& Test Europe Conf.}, 2018, pp. 241--244.

\bibitem{zhu2019sparse}
M.~Zhu, T.~Zhang, Z.~Gu, and Y.~Xie, ``Sparse tensor core: Algorithm and
  hardware co-design for vector-wise sparse neural networks on modern {GPUs},''
  in \emph{Proc. Int. Symp. Microarchitecture}, 2019, pp. 359--371.

\end{thebibliography}


\end{document}